\PassOptionsToPackage{table}{xcolor}
\documentclass[]{antgroup}
\usepackage{antgroup}

\usepackage{latexsym}

\usepackage[T1]{fontenc}

\usepackage[utf8]{inputenc}

\usepackage{amsmath}
\usepackage{amssymb}
\usepackage{booktabs}
\usepackage{array}
\usepackage{enumitem}
\usepackage{graphicx}
\usepackage{multirow}
\usepackage{xspace}
\usepackage{xcolor}
\usepackage{tikz}
\usepackage{pgfplots}
\usepackage{float}
\usepackage{placeins}
\usepackage{hyperref}
\usepackage{xurl}
\hypersetup{hidelinks}
\pgfplotsset{compat=1.18}
\usetikzlibrary{arrows.meta,calc,decorations.pathreplacing,positioning,shadows}

\usepackage{amsmath,amsfonts,bm}

\def\eqref#1{equation~\ref{#1}}

\def\1{\bm{1}}

\DeclareMathAlphabet{\mathsfit}{\encodingdefault}{\sfdefault}{m}{sl}
\SetMathAlphabet{\mathsfit}{bold}{\encodingdefault}{\sfdefault}{bx}{n}

\definecolor{pcdgreen}{HTML}{2F7D32}
\definecolor{pcdgray}{HTML}{5F6368}
\definecolor{pcdrow}{HTML}{F2F3F5}
\definecolor{figtwoblue}{HTML}{5FA0D6}
\definecolor{figtwoyellow}{HTML}{F3D582}
\definecolor{lladablue}{HTML}{4F7FB8}
\definecolor{lladaorange}{HTML}{F26B3A}
\newcommand{\best}[1]{\textbf{#1}}

\title{Reducing Pretraining-Generation Mismatch in Diffusion Language Models}

\author{\centering Xiaocheng Lu$^{2,1,*}$ \quad Huabin Liu$^{1,*}$ \quad Song Guo$^{2,\dagger}$ \quad Jianguo Li$^{1,\dagger}$}
\affiliation{\centering $^1$Inclusion AI \quad $^2$The Hong Kong University of Science and Technology}

\begin{document}
\maketitle
\begingroup
\renewcommand{\thefootnote}{}
\begin{NoHyper}
\footnotetext{\textsuperscript{*}Equal contribution. \textsuperscript{\ensuremath{\dagger}}Corresponding authors.\\
\hspace*{1.8em}This work was done during an internship at Inclusion AI.}
\end{NoHyper}
\endgroup

\begin{abstract}
Diffusion language models (dLLMs) generate text through iterative denoising, allowing multiple tokens to be predicted in parallel.
However, pretraining may mask tokens throughout a sequence, whereas prompt continuation conditions on an intact prefix.
This difference remains in conversion pipelines that denoise entire sequences during the stable stage.
We propose \emph{Prefix-Conditioned Diffusion} (PCD), which samples a boundary, preserves the prefix, and denoises the suffix.
The training recipe also applies autoregressive supervision to the prefix. We evaluate PCD in the stable stage of a warmup, stable, and decay conversion pipeline, with inference unchanged.
Matched experiments across model families show improvements in reasoning and coding over native diffusion training.
A matched continuation study further shows that the advantage persists after a shared decay stage.
In a separate reconstruction diagnostic, the full PCD recipe's advantage over native diffusion training reverses as more evaluation prefix tokens are masked.
Controlled experiments show lower suffix reconstruction loss with a clean training prefix, an intact evaluation prefix, and no autoregressive loss.
\end{abstract}

\section{Introduction}

Diffusion language models (dLLMs) generate text through iterative denoising, predicting multiple tokens in parallel \citep{nie2025large}.
For prompt continuation, they condition on an intact prefix. Autoregressive (AR) training reflects this structure by predicting each token from its preceding context \citep{vaswani2017attention,brown2020language}.
Diffusion pretraining over entire sequences, however, may mask tokens in this conditioning prefix.
This difference motivates testing whether preserving prefix context during pretraining improves prompt continuation.

\begin{figure}[!t]
  \centering
  \includegraphics[width=0.88\linewidth]{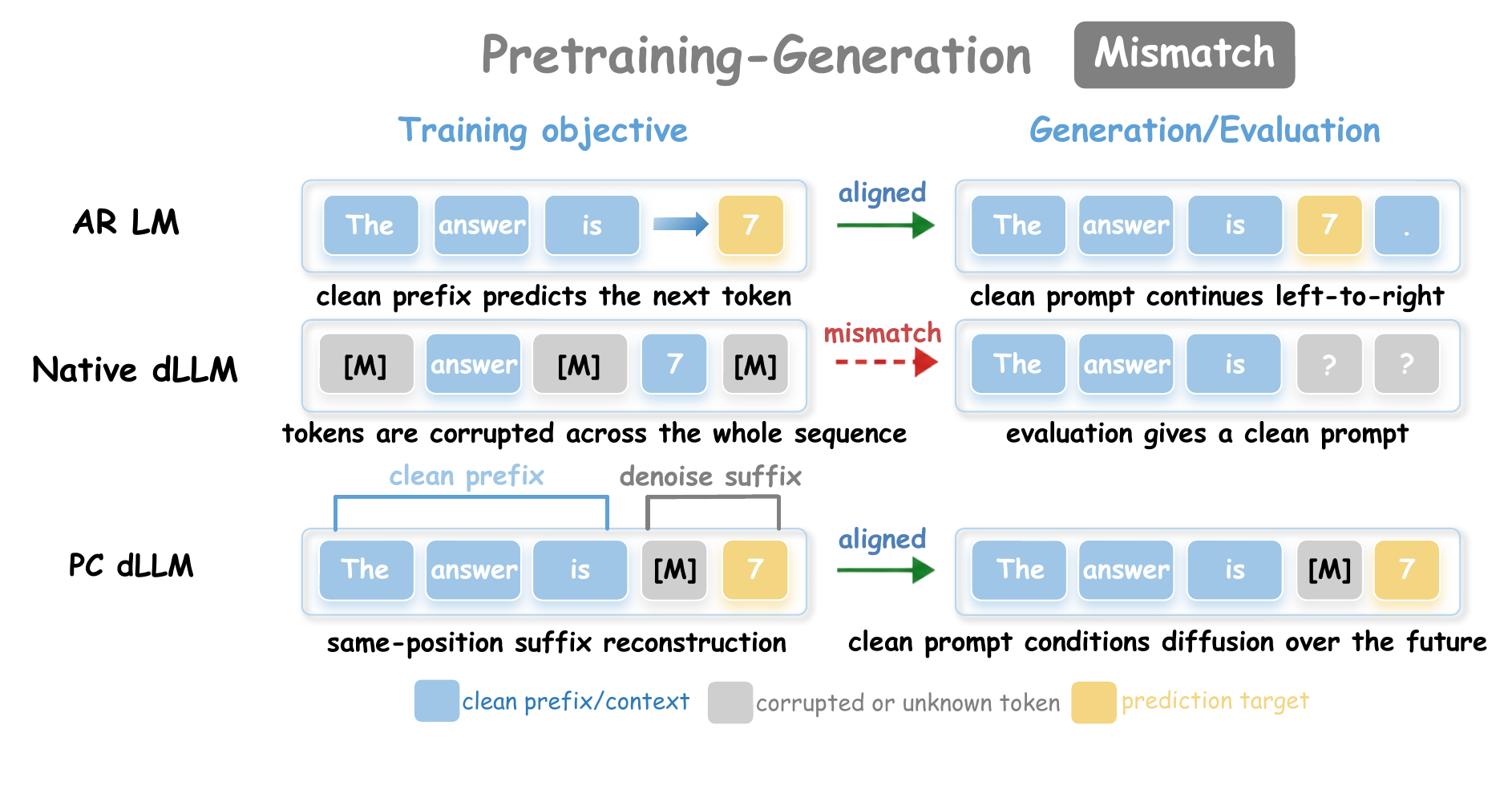}
  \caption{Motivation for Prefix-Conditioned Diffusion (PCD). Native diffusion pretraining can corrupt the prefix even though prompt continuation provides it intact at evaluation. The full PCD recipe applies an AR loss to the prefix and an MDM loss to the suffix. Its defining interface change is that suffix denoising conditions on a clean prefix.}
  \label{fig:motivation-mismatch}
\end{figure}

Existing block diffusion methods already condition on clean preceding blocks \citep{arriola2025block,jain2026adaptive}.
This conditioning structure need not be maintained throughout the conversion of an AR model into a dLLM.
For example, LLaDA2.0 uses a warmup, stable, and decay block-size schedule, analogous to WSD learning-rate scheduling \citep{wen2025understanding,bie2025llada2}.
Its stable stage denoises entire sequences before returning to block diffusion during decay.
We use this pipeline to test whether preserving a clean prefix during stable training improves downstream performance and whether the benefit persists after the same decay.

Our approach constructs the conditioning structure of prompt continuation directly from unlabeled text.
\emph{Prefix-Conditioned Diffusion} (PCD) samples a boundary within each sequence, preserves the prefix, and applies diffusion corruption only to the suffix (Figure~\ref{fig:motivation-mismatch}).
Sampling the boundary exposes the model to different amounts of visible context without requiring annotated prompt and response pairs.
The full recipe also applies causal AR supervision to the prefix, retaining the objective of predicting the next token alongside suffix denoising.
The objective does not require WSD. Here, we apply it during stable training while retaining the warmup, decay, and inference procedures.

Matched comparisons across two model families show reasoning and coding gains over native diffusion training.
On LLaDA2-Mini, the advantage persists after both checkpoints undergo the same decay procedure.
Separate public controls use block diffusion at a smaller budget. With attention and suffix supervision fixed and no AR loss, preserving the training prefix lowers reconstruction loss under evaluation with an intact prefix.
A different diagnostic shows that the full recipe's reconstruction advantage over native training reverses as more evaluation prefix tokens are masked.

Our contributions are threefold:
\begin{itemize}[leftmargin=1.5em,itemsep=1pt,topsep=2pt,parsep=0pt]
  \item \textbf{Prefix-conditioned training.} PCD partitions each sequence at a sampled boundary, preserving the prefix for suffix denoising to reduce the mismatch with prompt continuation. The full recipe combines causal prefix AR supervision with same-position suffix reconstruction.
  \item \textbf{Task gains and persistence through decay.} Matched comparisons across two model families show task gains for the complete recipe. In the LLaDA2-Mini comparison, an average advantage remains after both checkpoints undergo the same decay procedure.
  \item \textbf{Controlled evidence for prefix visibility.} A separate public control fixes attention and suffix supervision without AR loss. Clean training prefixes lower reconstruction loss under clean-prefix evaluation, isolating a visibility benefit within the tested block-causal setting.
\end{itemize}

\section{Related Work}
\label{sec:related-work}

\paragraph{Diffusion language models.}
Diffusion models generate data through iterative denoising \citep{sohl-dickstein2015deep,ho2020denoising,song2021score}, with discrete, categorical, and continuous formulations for language \citep{hoogeboom2021argmax,austin-etal-2021-structured,li-etal-2022-diffusion-lm,gong2023diffuseq}.
Discrete dLLMs improve objectives and sampling \citep{lou2024discrete,sahoo2024simple}, while LLaDA and Dream demonstrate large-scale pretraining and supervised fine-tuning \citep{nie2025large,ye2025dream}. Earlier continuous-diffusion work reports prefix-masking trade-offs across conditioning settings \citep{locicerovaina2023halting}.
WSD supports branching and learning-rate decay in language-model pretraining \citep{wen2025understanding}; LLaDA2.0 uses a WSD-like recipe to preserve AR knowledge, train with full-sequence diffusion, and decay to block-diffusion checkpoints \citep{bie2025llada2}.
Nemotron-Labs-Diffusion jointly trains AR and diffusion objectives with a strictly causal clean stream and a noisy stream, supporting AR, diffusion, and self-speculative decoding \citep{fu2026nemotronlabsdiffusion}.
Joint AR/denoising training is thus established and does not itself distinguish PCD.

\paragraph{Block diffusion.}
Block diffusion combines autoregression across blocks with denoising within blocks for arbitrary-length generation; BD3-LM trains noisy blocks conditioned on clean preceding blocks \citep{arriola2025block}.
Blockwise SFT aligns one active response block with semi-autoregressive inference by freezing preceding response blocks and hiding future ones \citep{sun2025blockwise}; prompt-infilling work instead extends masking over prompts during SFT \citep{fujinuma2026unlocking}.
Adaptive Block Diffusion samples prefix-window configurations and denoises an active window given clean prefix context \citep{jain2026adaptive}, while training-free analyses expose a left-to-right generation bias in block-diffusion checkpoints \citep{hou2026rethinking}.
Decoding efficiency is also addressed through confidence-based horizon adaptation \citep{lu2026pace} and on-policy transition distillation \citep{lu2026optd}.
PCD uses a single clean-prefix/noisy-suffix sequence, rather than Nemotron-Labs-Diffusion's dual streams. Its sampled prefix and remaining suffix form a restricted case of Adaptive Block Diffusion's prefix-window configurations; the full PCD recipe also includes prefix AR supervision.
Our WSD comparisons ask whether applying PCD during stable training leaves a task advantage after the same decay; the public intervention tests prefix visibility at fixed attention and supervision.
Without matched training comparisons, our results do not rank these methods.

\section{Method}

\begin{figure*}[!t]
  \centering
  \begin{tikzpicture}
    \node[inner sep=0,anchor=south west] (art) at (0,0) {\includegraphics[width=0.98\textwidth]{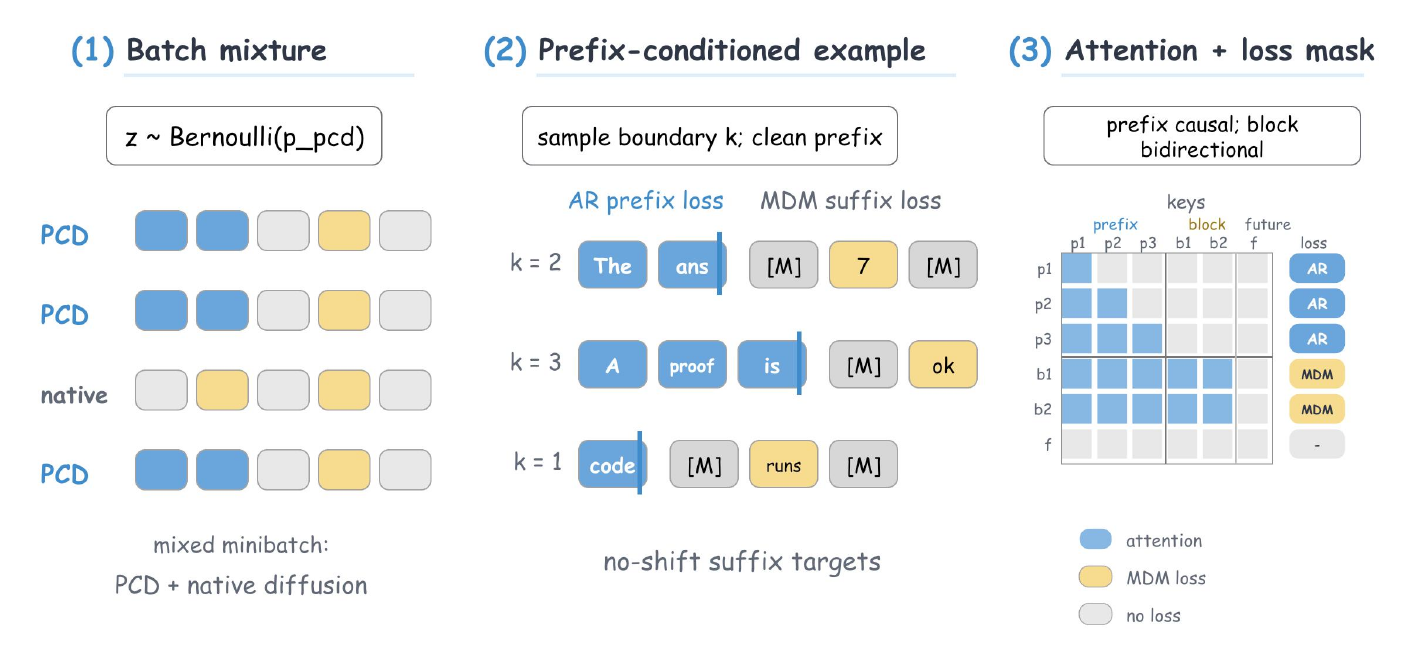}};
    \begin{scope}[x={(art.south east)},y={(art.north west)}]
      \fill[white] (0.911,0.462) rectangle (0.955,0.513);
      \fill[black!9,rounded corners=1.5pt] (0.913,0.465) rectangle (0.953,0.509);
      \node[inner sep=0,font=\sffamily\tiny,text=black!65] at (0.933,0.487) {--};
    \end{scope}
  \end{tikzpicture}
  \caption{PCD training. Panel~(3) shows block-causal attention for prefix and active-block queries; future queries are omitted. Full-scale suffix attention is bidirectional. AR logits end at $k-1$ (none for $k=1$); MDM loss uses masked suffix positions only.}
  \label{fig:pcd-training-detail}
\end{figure*}
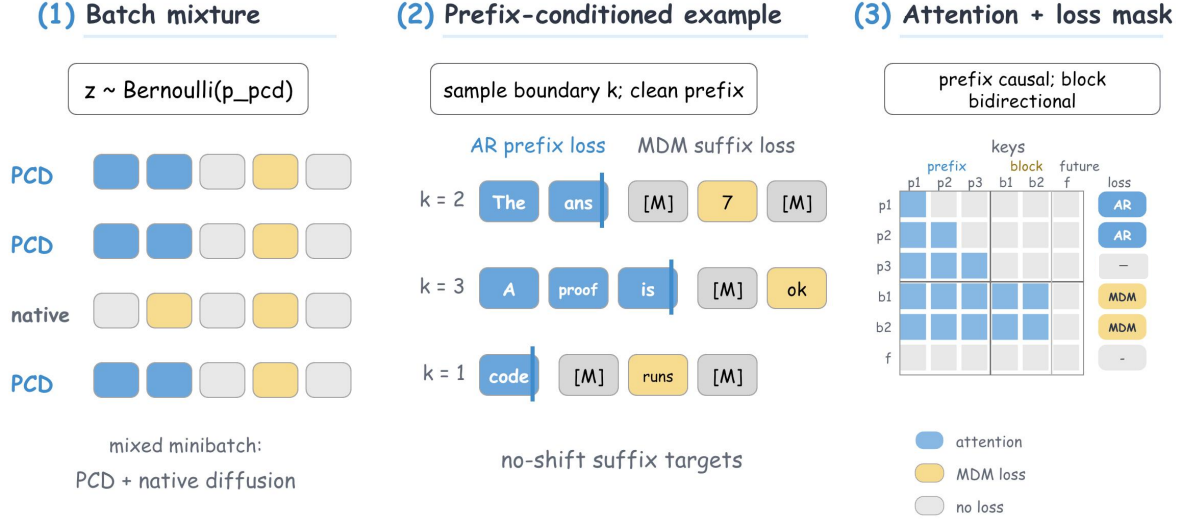

PCD preserves a sampled prefix during denoising, independently of the training schedule. We define its corruption rule, attention pattern, and joint objective before describing the training procedure and its application to WSD (Figure~\ref{fig:pcd-training-detail}).

\subsection{Prefix-Preserving Denoising}
\label{sec:prefix-objective}

Let $x=(x_1,\ldots,x_T)$ be a sequence and $m_i=1$ mark a prediction target. Masked diffusion replaces selected tokens with $\texttt{[MASK]}$ and leaves the others unchanged, producing $\tilde x$. Native training allows targets at any position and minimizes
\begin{equation}
  \mathcal L_{\mathrm{native}}=
  -\mathbb E_{x,m}\frac{\sum_{i:m_i=1}\log p_\theta(x_i\mid\tilde x,m)}{\max(\sum_i m_i,1)}.
  \label{eq:native-loss}
\end{equation}
Conditioning denotes inputs accessible under the attention mask. This is the per-example normalized cross-entropy used here, not a general diffusion likelihood bound.

PCD samples $\rho\sim q(\rho)$ and sets $k=\min(T-1,\max(1,\lfloor\rho T\rfloor))$, keeping both regions nonempty. Sampling the boundary varies the amount of conditioning context without requiring annotated prompt/response pairs. With $m_i^{\mathrm{pcd}}=\mathbb I[i>k]m_i$, its input is
\begin{equation}
  \tilde x_i^{\mathrm{pcd}}=
  \begin{cases}
    \texttt{[MASK]},&m_i^{\mathrm{pcd}}=1,\\
    x_i,&\text{otherwise}.
  \end{cases}
\end{equation}
Only masked suffix positions receive denoising targets. Noise sampling is a recipe choice; the public controls draw a masking rate and suffix masks, ensuring at least one target. These settings are not assigned to the original full-scale runs.

\subsection{Attention and Joint Supervision}
\label{sec:attention-supervision}

Let $A_{ij}=1$ allow query position $i$ to read key position $j$. PCD uses
\begin{equation}
 A_{ij}=\begin{cases}
 \mathbb I[j\le i],&i\le k,\\
 1,&i>k,\ j\le k,\\
 S_{ij},&i>k,\ j>k.
 \end{cases}
 \label{eq:pcd-attention}
\end{equation}
The causal prefix cannot read suffix tokens, preventing leakage into its AR predictions. Every suffix query can read the intact prefix. Full-scale stable training uses bidirectional suffix attention, $S_{ij}=1$. The public controls instead use $S_{ij}=\mathbb I[b(j)\le b(i)]$, where $b(i)=\lfloor(i-k-1)/B\rfloor$ for $i>k$ and block size $B$. Thus their suffix blocks begin at $k+1$ and attend to current and preceding blocks.

The complete recipe adds AR supervision to train on the preserved prefix. Its losses are
\begin{equation}
\begin{aligned}
  \ell_{\mathrm{suf}}&=-\frac{1}{N_m}\sum_{i:m_i^{\mathrm{pcd}}=1}
    \log p_\theta(x_i\mid\tilde x^{\mathrm{pcd}},m^{\mathrm{pcd}},k),\\
  \ell_{\mathrm{ar}}&=-\frac{1}{N_{\mathrm{ar}}}\sum_{i=2}^{k}
    \log p_\theta(x_i\mid x_{<i}),\\
  \mathcal L_{\mathrm{pcd}}&=\mathbb E_{x,\rho,m}
    [\lambda_{\mathrm{mdm}}\ell_{\mathrm{suf}}+\lambda_{\mathrm{ar}}\ell_{\mathrm{ar}}],
\end{aligned}
\label{eq:intra-pcd}
\end{equation}
where $N_m=\max(\sum_i m_i^{\mathrm{pcd}},1)$ and $N_{\mathrm{ar}}=\max(k-1,1)$. Prefix logits at positions $1,\ldots,k-1$ predict $x_2,\ldots,x_k$; position $k$ has no AR target. Suffix logits reconstruct tokens at the same positions (``no-shift''). If $k=1$, the AR loss is zero. The boundary $k$ specifies attention and supervision, not a learned input.

Each region is normalized by its own target count before weighting, so a longer region does not dominate merely by having more targets. Unit weights assign equal coefficients to region averages, not individual tokens. The full-scale recipe uses $\lambda_{\mathrm{ar}}=\lambda_{\mathrm{mdm}}=1$ unless varied in a sweep. Setting $\lambda_{\mathrm{ar}}=0$ removes prefix supervision while retaining the same PCD inputs, attention, and suffix targets; the downstream benefit of the AR term remains an empirical question.

\subsection{Training Recipe and WSD Evaluation Setting}
\label{sec:stable-recipe}

The joint objective provides \emph{intra-sample} mixing of AR and denoising. Optional \emph{inter-sample} mixing selects PCD with probability $p_{\mathrm{pcd}}$ and native diffusion otherwise:
\begin{equation}
  \mathcal L=p_{\mathrm{pcd}}\mathcal L_{\mathrm{pcd}}
    +(1-p_{\mathrm{pcd}})\mathcal L_{\mathrm{native}}.
\end{equation}
For each example, select the objective branch; for PCD, sample the boundary, corrupt the suffix, and construct Equation~\ref{eq:pcd-attention}. Compute the selected branch's normalized losses, average over examples, and update the model. The prefix distribution, mixing probability, and AR weight are independent choices. At prefix range $[1,1]$, clamping gives $k=T-1$, leaving one denoising token; we therefore call its mixture with native training \emph{AR-dominant/native mixing}.

WSD is our evaluation setting, not a requirement of PCD. We apply PCD during stable training after a shared AR-to-block-diffusion warmup. Matched decay then tests whether the task advantage persists under the same subsequent training and inference. PCD requires no AR decoding, verification, or self-speculation at inference.

\section{Experiments}
\label{sec:experiments}

\begin{table}[!t]
  {\small
  \setlength{\abovecaptionskip}{0pt}
  \setlength{\belowcaptionskip}{10pt}
  \caption{Main results with external scale references. Avg.6 averages the six displayed tasks. External rows contextualize model scale; the LLaDA2-Mini rows provide the controlled comparison, with 400B stable tokens and a matched 50B-token decay. Shading marks PCD; bold marks the best LLaDA2-Mini score in each metric. Table~\ref{tab:main-coverage} reports the remaining six tasks.}
  \label{tab:main-results}
  }
  \centering
  \renewcommand{\arraystretch}{1.15}
  \small
  \setlength{\tabcolsep}{3pt}
  \resizebox{\textwidth}{!}{%
  \begin{tabular}{llrrrrrrr}
    \toprule
    Model & Params & Avg.6 & GSM8K & MATH & BBH & KorBench & MBPP & \shortstack{HumanEval-\\FIM} \\
    \midrule
    \multicolumn{9}{l}{\textit{External dLLM references}} \\
    LLaDA base & 8B & 50.12 & 73.62 & 31.48 & 47.18 & 32.72 & 40.60 & 75.12 \\
    LLaDA-MoE base & 7B-A1B & 51.94 & 66.41 & 36.10 & 52.71 & 31.20 & 52.40 & 72.80 \\
    Dream base & 7B & 57.48 & 81.58 & 41.50 & 53.38 & 37.68 & 53.20 & 77.54 \\
    \midrule
    \multicolumn{9}{l}{\textit{External AR scale references}} \\
    LLaMA3-8B base & 8B & 48.50 & 58.07 & 17.74 & 58.83 & 33.04 & 48.20 & 75.12 \\
    Qwen2.5-7B base & 7B & 61.87 & 86.28 & 51.10 & 56.79 & 35.92 & 62.40 & 78.70 \\
    \midrule
    \multicolumn{9}{l}{\textit{Representative matched LLaDA2-Mini comparison}} \\
    LLaDA2-Mini native stable & 16B-A1B & 60.30 & 75.82 & 56.38 & 66.38 & 33.12 & 59.40 & 70.67 \\
    LLaDA2-Mini native decay & 16B-A1B & 62.87 & 80.82 & 56.92 & 68.96 & \best{37.36} & 59.40 & 73.77 \\
    \rowcolor{pcdrow}
    LLaDA2-Mini \textbf{PCD stable} & 16B-A1B & 62.86 & 80.44 & 57.64 & 68.40 & 35.84 & 60.20 & \best{74.64} \\
    \rowcolor{pcdrow}
    LLaDA2-Mini \textbf{PCD decay} & 16B-A1B & \best{64.54} & \best{83.47} & \best{59.52} & \best{69.63} & 37.28 & \best{63.20} & 74.15 \\
    \bottomrule
  \end{tabular}
  }
\end{table}

\begin{table}[!t]
  {\small
  \setlength{\abovecaptionskip}{0pt}
  \setlength{\belowcaptionskip}{10pt}
  \caption{Remaining six tasks in the matched LLaDA2-Mini comparison. Together with Table~\ref{tab:main-results}, these rows display all 12 task scores; Avg.12 averages the complete suite. Bold marks the better value within each matched budget pair.}
  \label{tab:main-coverage}
  }
  \centering
  \renewcommand{\arraystretch}{1.15}
  \small
  \setlength{\tabcolsep}{3pt}
  \resizebox{\textwidth}{!}{%
  \begin{tabular}{lrrrrrrr}
    \toprule
    Checkpoint & Avg.12 & NQ & TriviaQA & MMLU-Pro & OmniMath & LCBench & AutoLogi \\
    \midrule
    Native stable (400B) & 47.26 & 25.84 & 54.27 & 44.74 & 19.70 & 27.54 & 33.27 \\
    \rowcolor{pcdrow}
    PCD stable (400B) & \best{49.74} & \best{29.17} & \best{57.46} & \best{46.08} & \best{20.90} & \best{30.72} & \best{35.37} \\
    \midrule
    Native decay (+50B) & 49.30 & 27.67 & 55.80 & 43.94 & 18.50 & 29.56 & \best{38.95} \\
    \rowcolor{pcdrow}
    PCD decay (+50B) & \best{50.94} & \best{29.53} & \best{57.99} & \best{46.10} & \best{20.80} & \best{31.41} & 38.21 \\
    \bottomrule
  \end{tabular}}
\end{table}

\subsection{Setup}

We evaluate LLaDA2-Mini and Qwen-1.7B checkpoints across reasoning, coding, math, and knowledge benchmarks. Within each native/PCD pair, no-shift SGLang settings, 8k context, prompts, initialization, data, and token budget are matched.
The controlled baseline is the same-family native stable objective, which corrupts the sequence without preserving a clean prefix.
Public dLLMs such as LLaDA, LLaDA-MoE, and Dream~\citep{nie2025large,zhu2025lladamoe,ye2025dream}, together with AR checkpoints, provide scale references rather than controlled comparisons.
We use the LLaDA2-Mini 400B stable pair as the primary recipe comparison, matched +50B decay as a continuation check, and Qwen-1.7B at 50B tokens as a second-backbone task study.
These comparisons evaluate the complete PCD objective relative to native stable training, rather than isolating each ingredient.
For PCD examples, suffix targets are unshifted and losses use per-region target normalization. The full-scale runs use proprietary continued-pretraining data. Their original records do not recover the main prefix distribution, PCD/native mixing probability, training sequence length, global batch, optimizer schedule, or complete decoding and task-scoring configurations. Settings explicitly reported for a sweep or public control apply only to those experiments.
Separate public-data reconstruction controls use Qwen3-1.7B-Base and FineWeb-Edu at a smaller budget and with block-causal suffix attention.
For seeds 42--44, four recipes branch from one 20M-token warmup for 50M stable tokens: native MDM, PCD, clean-prefix/no-AR, and full-sequence AR plus native MDM in separate views.
A follow-up with seeds 52--54 changes only training-prefix corruption and uses a fresh 2M-token holdout.
Both public studies pair evaluator draws across checkpoints and measure masked-token reconstruction cross-entropy. Their 50M-token stable budgets are distinct from the 50B-token Qwen task study; they test reconstruction behavior rather than reproduce its task gains.

\subsection{Benchmarks}

The LLaDA2-Mini suite covers knowledge/QA (NQ~\citep{kwiatkowski-etal-2019-natural}, TriviaQA~\citep{joshi-etal-2017-triviaqa}, MMLU-Pro~\citep{wang2024mmlupro}), math (OmniMath~\citep{gao2024omnimathuniversalolympiadlevel}, MATH~\citep{hendrycks2021math}, GSM8K~\citep{cobbe2021trainingverifierssolvemath}), code (MBPP~\citep{austin2021program}, HumanEval-FIM, LiveCodeBench (LCBench)~\citep{jain2024livecodebenchholisticcontaminationfree}), and general reasoning (BBH~\citep{suzgun2022challengingbigbenchtaskschainofthought}, KorBench~\citep{ma2024korbench}, AutoLogi~\citep{zhu2025autologi}).
We report an infilling variant of HumanEval~\citep{chen2021evaluatinglargelanguagemodels} as HumanEval-FIM. The Qwen-1.7B suite uses the six benchmarks listed in its tables, including standard HumanEval rather than HumanEval-FIM.
Task scores use a 0--100 scale (higher is better). Avg.6 and Avg.12 are unweighted means of the specified task scores; gains are absolute score-point differences. Chat-SFT averages five tasks. Public reconstruction controls report masked-suffix cross-entropy in nats (lower is better).

\begin{figure}[!t]
  \centering
  \begin{minipage}[t]{0.48\textwidth}
    \vspace{0pt}
    \centering
    \begin{tikzpicture}[baseline=(current bounding box.north)]
      \begin{axis}[
        width=\linewidth,
        height=4.2cm,
        title={Matched budget gaps},
        title style={font=\small\bfseries},
        xlabel={Continued-pretraining budget},
        ylabel={6-bench average},
        xmin=90, xmax=455,
        ymin=58.5, ymax=65.2,
        xtick={100,200,300,400,450},
        xticklabels={100B,200B,300B,400B,+50B},
        x tick label style={font=\scriptsize},
        ytick={59,60,61,62,63,64,65},
        grid=major,
        major grid style={gray!22},
        axis x line*=bottom,
        axis y line*=left,
        axis line style={black!75},
        tick label style={font=\footnotesize},
        label style={font=\footnotesize},
        legend style={font=\scriptsize,draw=gray!35,fill=white,fill opacity=0.92,text opacity=1,at={(0.03,0.97)},anchor=north west},
        legend cell align={left},
      ]
        \path[fill=gray!8,draw=none]
          (axis cs:400,58.5) rectangle (axis cs:450,65.2);
        \addplot[pcdgray!45,densely dashed,line width=0.4pt,forget plot] coordinates {
          (400,58.5) (400,65.2)
        };
        \addplot[draw=none,fill=lladaorange!14,forget plot] coordinates {
          (100,61.13) (200,61.97) (300,62.14) (400,62.86) (450,64.54)
          (450,62.87) (400,60.30) (300,59.54) (200,60.64) (100,60.14)
        } \closedcycle;
        \addplot+[thick,mark=square*,mark size=1.6pt,color=lladablue,mark options={fill=lladablue}] coordinates {
          (100,60.14) (200,60.64) (300,59.54) (400,60.30) (450,62.87)
        };
        \addlegendentry{Native dLLM}
        \addplot+[very thick,mark=*,mark size=1.8pt,color=lladaorange,mark options={fill=lladaorange}] coordinates {
          (100,61.13) (200,61.97) (300,62.14) (400,62.86) (450,64.54)
        };
        \addlegendentry{PCD}
        \node[anchor=west,font=\scriptsize\bfseries,text=lladaorange!80!black] at (axis cs:102,60.72) {+0.99};
        \node[font=\scriptsize\bfseries,text=lladaorange!80!black] at (axis cs:200,61.28) {+1.33};
        \node[font=\scriptsize\bfseries,text=lladaorange!80!black] at (axis cs:300,60.92) {+2.60};
        \node[font=\scriptsize\bfseries,text=lladaorange!80!black] at (axis cs:400,61.58) {+2.56};
        \node[anchor=east,font=\scriptsize\bfseries,text=lladaorange!80!black] at (axis cs:448,63.68) {+1.67};
        \node[font=\tiny\bfseries,text=pcdgray,anchor=north] at (axis cs:425,65.12) {decay};
      \end{axis}
    \end{tikzpicture}
    \caption{Matched budget gaps. PCD stays above native dLLM across stable budgets and matched +50B decay.}
    \label{fig:budget-gaps}
  \end{minipage}\hfill
  \begin{minipage}[t]{0.48\textwidth}
    \vspace{0pt}
    \centering
    \begin{tikzpicture}[baseline=(current bounding box.north)]
      \begin{axis}[
        width=\linewidth,
        height=4.2cm,
        title={Gains across four domains},
        title style={font=\small\bfseries},
        ylabel={Domain average},
        xlabel={Benchmark domain},
        ymin=38, ymax=58,
        ytick={40,45,50,55},
        symbolic x coords={QA,General,Math,Code},
        xtick=data,
        x tick label style={font=\scriptsize},
        yticklabel style={font=\footnotesize},
        label style={font=\footnotesize},
        grid=major,
        major grid style={gray!18},
        axis x line*=bottom,
        axis y line*=left,
        axis line style={black!75},
        enlarge x limits=0.12,
        legend style={font=\scriptsize,draw=none,at={(0.02,0.98)},anchor=north west},
      ]
        \addplot+[only marks,mark=square*,mark size=2.5pt,color=lladablue] coordinates {
          (QA,41.62) (General,44.26) (Math,50.63) (Code,52.54)
        };
        \addlegendentry{Native}
        \addplot+[only marks,mark=*,mark size=2.5pt,color=lladaorange] coordinates {
          (QA,44.24) (General,46.54) (Math,52.99) (Code,55.19)
        };
        \addlegendentry{PCD}
        \node[anchor=south,font=\scriptsize\bfseries,text=lladaorange!80!black] at (axis cs:QA,44.24) {+2.62};
        \node[anchor=south,font=\scriptsize\bfseries,text=lladaorange!80!black] at (axis cs:General,46.54) {+2.28};
        \node[anchor=south,font=\scriptsize\bfseries,text=lladaorange!80!black] at (axis cs:Math,52.99) {+2.36};
        \node[anchor=south,font=\scriptsize\bfseries,text=lladaorange!80!black] at (axis cs:Code,55.19) {+2.65};
      \end{axis}
    \end{tikzpicture}
    \caption{12-benchmark domain averages at the final stable stage. Orange labels report PCD gains.}
    \label{fig:domain-averages}
  \end{minipage}
\end{figure}

\subsection{Main Results}

Table~\ref{tab:main-results} reports external dLLM and AR references alongside the matched LLaDA2-Mini stable and decay results.
The stable pair tests the objective replacement; the +50B pair tests whether adapting both parents through the same decay removes the resulting advantage.
At the matched 400B stable budget, PCD improves the native dLLM stable baseline by +2.56 Avg.6 points.
Figure~\ref{fig:budget-gaps} shows that the advantage persists along the paired training trajectories: PCD improves over the same-family native baseline by +0.99 Avg.6 at 100B, +1.33 at 200B, +2.60 at 300B, and +2.56 at 400B.
At 400B, PCD improves all 12 tasks in the wider suite, raising Avg.12 from 47.26 to 49.74 (+2.48); the gain spans knowledge/QA, math, code, and general reasoning. Tables~\ref{tab:main-results} and~\ref{tab:main-coverage} report all 12 tasks.

After matched +50B decay, PCD retains +1.67 Avg.6 and +1.64 Avg.12 points, improving 10/12 tasks.
The exceptions are KorBench ($-0.08$) and AutoLogi ($-0.74$); these are descriptive differences on the reported task scores.
These scores describe one paired full-scale trajectory; Section~\ref{sec:analysis} discusses uncertainty and attribution.
Chat-SFT transfer and AR-decoding compatibility are examined in Section~\ref{sec:analysis} (Figures~\ref{fig:chat-sft-transfer} and~\ref{fig:ar-interface-sanity}).

\begin{table}[!t]
  {\small
  \setlength{\abovecaptionskip}{0pt}
  \setlength{\belowcaptionskip}{10pt}
  \caption{Qwen-1.7B objective variants at 50B continued pretraining. HE denotes standard HumanEval. All use matched initialization, data, budget, and evaluation. Bold marks the best value in this controlled set; the close Avg.6 values do not identify a unique causal component.}
  \label{tab:qwen-mechanism-summary}
  }
  \centering
  \renewcommand{\arraystretch}{1.15}
  \small
  \setlength{\tabcolsep}{2.5pt}
  \resizebox{\textwidth}{!}{%
  \begin{tabular}{lccrrrrrrr}
    \toprule
    Setting & $p_{\mathrm{pcd}}$ & Prefix range & Avg.6 & HE & GSM8K & MATH & BBH & KorBench & MBPP \\
    \midrule
    Native baseline & 0.00 & -- & 34.28 & 25.00 & 59.14 & 26.84 & 39.72 & 20.96 & 34.00 \\
    \rowcolor{pcdrow}
    Mixed PCD & 0.50 & [0.0, 1.0] & 38.94 & 28.05 & 66.94 & 33.36 & 45.87 & 22.00 & 37.40 \\
    \rowcolor{pcdrow}
    Intra-sample only & 1.00 & [0.0, 1.0] & \best{39.14} & 25.61 & \best{67.70} & \best{33.68} & \best{46.20} & 21.84 & \best{39.80} \\
    \rowcolor{pcdrow}
    AR-dominant/native mix & 0.50 & [1.0, 1.0] & 38.98 & \best{29.88} & 62.93 & \best{33.68} & 45.84 & \best{23.12} & 38.40 \\
    \bottomrule
  \end{tabular}
  }
\end{table}

\subsection{Recipe Controls and Prefix Visibility}

\paragraph{Warmup and continuation.}
Under the tested Qwen recipe, stable PCD performs much better after AR-to-BD warmup than when initialized directly from an AR checkpoint. We therefore retain that warmup in the studied pipeline. Without a direct AR-to-native control, this diagnostic cannot identify a PCD-specific dependence on warmup.
Longer PCD-only continuations lack corresponding native runs and do not extend the matched decay comparison.

\begin{figure}[!t]
  \centering
  \begin{minipage}[t]{0.48\textwidth}
    \vspace{0pt}
    \centering
    \begin{tikzpicture}[baseline=(current bounding box.north)]
      \begin{axis}[
        width=\linewidth,
        height=4.2cm,
        title={Qwen recipe sweep},
        title style={font=\small\bfseries},
        xmin=-0.03, xmax=1.03,
        ymin=0, ymax=8.2,
        xtick={0,0.25,0.50,0.75,1.0},
        xlabel={Ratio / probability},
        ylabel={Avg.6 gain},
        grid=both,
        major grid style={gray!18},
        axis x line*=bottom,
        axis y line*=left,
        axis line style={black!75},
        tick label style={font=\footnotesize},
        label style={font=\footnotesize},
        legend style={font=\scriptsize,draw=none,at={(0.04,0.97)},anchor=north west,fill=white,fill opacity=0.85,text opacity=1},
      ]
        \addplot+[pcdgreen,thick,mark=*,mark options={fill=pcdgreen}] coordinates {
          (0.25,2.23)
          (0.50,5.61)
          (0.75,7.04)
          (1.00,4.86)
        };
        \addlegendentry{Intra $[0,r]$}
        \addplot+[pcdgray,thick,mark=square*,mark options={fill=pcdgray}] coordinates {
          (0.00,0.00)
          (0.25,4.57)
          (0.50,4.70)
          (0.75,2.06)
          (1.00,3.60)
        };
    \addlegendentry{Inter $p$, $[1,1]$}
        \node[anchor=south,font=\tiny\bfseries,text=pcdgreen] at (axis cs:0.75,7.04) {+7.0};
        \node[anchor=north,font=\scriptsize\bfseries,text=pcdgray] at (axis cs:0.50,4.70) {+4.7};
      \end{axis}
    \end{tikzpicture}
    \caption{Qwen-1.7B 50B-token Avg.6 gains. Intra varies $r$; Inter varies $p$ at the AR-dominant prefix endpoint $[1,1]$. Table~\ref{tab:qwen-mechanism-summary} compares the representative recipes.}
    \label{fig:qwen-ratio-sweep}
  \end{minipage}\hfill
  \begin{minipage}[t]{0.48\textwidth}
    \vspace{0pt}
    \centering
    \begin{tikzpicture}[baseline=(current bounding box.north)]
      \begin{axis}[
        width=\linewidth,
        height=4.2cm,
        title={Prefix-AR weight},
        title style={font=\small\bfseries},
        xlabel={Prefix-AR weight $\lambda_{\mathrm{ar}}$},
        ylabel={Avg.6 gain},
        xmin=0.18, xmax=1.07,
        ymin=0, ymax=8.2,
        xtick={0.25,0.50,0.75,1.0},
        ytick={0,2,4,6,8},
        grid=major,
        major grid style={gray!18},
        axis x line*=bottom,
        axis y line*=left,
        axis line style={black!75},
        tick label style={font=\footnotesize},
        label style={font=\footnotesize},
        legend style={font=\scriptsize,draw=none,at={(0.04,0.97)},anchor=north west,fill=white},
      ]
        \addplot+[lladaorange,thick,mark=diamond*,mark options={fill=lladaorange}] coordinates {
          (0.25,4.92) (0.50,5.74) (0.75,6.07) (1.00,4.66)
        };
        \addlegendentry{Mixed PCD}
        \node[anchor=south,font=\scriptsize\bfseries,text=lladaorange!80!black] at (axis cs:0.75,6.07) {+6.07};
      \end{axis}
    \end{tikzpicture}
    \caption{Qwen-1.7B 50B-token Avg.6 gains over native training. Mixed PCD fixes $p_{\mathrm{pcd}}=0.5$, prefix range $[0,1]$, and $\lambda_{\mathrm{mdm}}=1$; only $\lambda_{\mathrm{ar}}$ varies.}
    \label{fig:qwen-ar-weight}
  \end{minipage}
\end{figure}

\paragraph{Intra- vs. inter-sample mixing.}
Under matched Qwen-1.7B settings, intra-sample-only PCD gives a +4.86 Avg.6 gain; the AR-dominant/native mixture (+4.70) and mixed PCD (+4.66) yield similar reported averages (Table~\ref{tab:qwen-mechanism-summary}).
All three variants exceed native training; Section~\ref{sec:analysis} discusses their attribution limits.
The finer ratio/probability sweep in Figure~\ref{fig:qwen-ratio-sweep} shows that the strongest reported prefix range is $[0,0.75]$. Its numerical maxima describe the completed sweep, rather than a recipe selected by an independently documented validation rule.

\paragraph{Objective design.}
At 100B tokens, same-position PCD suffix targets achieve 61.13 Avg.6 versus 54.43 for shifted targets; mismatching training and evaluation shifts reduces the score to 0.64. In the Qwen mixed recipe, all tested prefix-AR weights from 0.25 to 1.00 improve over the native baseline; weight 0.75 reaches 40.35 Avg.6 (Figure~\ref{fig:qwen-ar-weight}). This sweep measures weight sensitivity; a matched zero-AR task comparison remains absent.

Clean-prefix/no-AR yields the lowest reconstruction loss in the public four-recipe screen. We test prefix visibility directly below by fixing attention and suffix supervision.

\begin{table}[!t]
  {\small
  \setlength{\abovecaptionskip}{0pt}
  \setlength{\belowcaptionskip}{10pt}
  \caption{Single-factor training-prefix intervention on Qwen3-1.7B. Values are per-example normalized masked-suffix cross-entropy in nats under the same clean-prefix evaluation interface (lower is better). Each row averages 976 blocks and three paired evaluator seeds; SD is across training seeds sharing one warmup. This intervention uses a fresh holdout and FP32 loss reduction; its absolute losses are not directly comparable with the earlier four-recipe screen.}
  \label{tab:matched-prefix-control}
  }
  \centering
  \renewcommand{\arraystretch}{1.15}
  \small
  \begin{tabular}{lrrr}
    \toprule
    Training seed & Clean-prefix training & Corrupted-prefix training & $\Delta$ (clean minus corrupted) \\
    \midrule
    52 & 3.9492 & 4.0173 & $-0.0681$ \\
    53 & 3.9514 & 4.0200 & $-0.0686$ \\
    54 & 3.9497 & 4.0205 & $-0.0708$ \\
    \midrule
    Mean $\pm$ SD & $3.9501\pm0.0012$ & $4.0193\pm0.0017$ & $-0.0692\pm0.0015$ \\
    \bottomrule
  \end{tabular}
\end{table}

\paragraph{Matched training-prefix intervention.}
We pair hybrid attention, suffix boundaries, masks, targets, normalization, data order, and optimization, with no AR loss; only training-prefix corruption differs.
The corrupted-prefix arm therefore retains the clean arm's attention and suffix-only supervision, unlike native MDM.
Both arms are evaluated with a clean prefix on the same fresh holdout.
Across three new training seeds sharing the warmup parent, the clean-minus-corrupted difference is $-0.0692$ nats (paired training-seed 95\% interval $[-0.0728,-0.0655]$); every seed improves, exceeding the prespecified $0.02$ reduction threshold.
Table~\ref{tab:matched-prefix-control} reports the individual seeds. Each branch uses 382 updates of 64 sequences of length 2,048, with $\rho\sim U[0,0.75]$, suffix block size 128, and masking rate $t\sim U[0.001,1]$. Prefix masking in the corrupted arm is independently sampled at the same $t$. Training uses AdamW with peak learning rate $2\times10^{-5}$, $\beta=(0.9,0.95)$, weight decay 0.1, gradient clipping at 1.0, and 20 warmup updates followed by cosine decay to 0.1 of the peak rate. The fresh holdout excludes historical document IDs and exact text hashes, but does not rule out semantic near-duplicates. The fixed success rule also requires every seed difference to be negative and the confidence interval upper bound to be below zero; both conditions hold.
This isolates a training-prefix benefit under the fixed block-causal reconstruction interface, not a decomposition of the full-scale task gains.

Native-to-PCD comparisons combine attention, targets, and conditioning, whereas the matched pair isolates training-prefix visibility under one reconstruction interface.

\FloatBarrier

\section{Analysis}
\label{sec:analysis}

\begin{figure}[!t]
  \centering
  \includegraphics[width=0.84\textwidth]{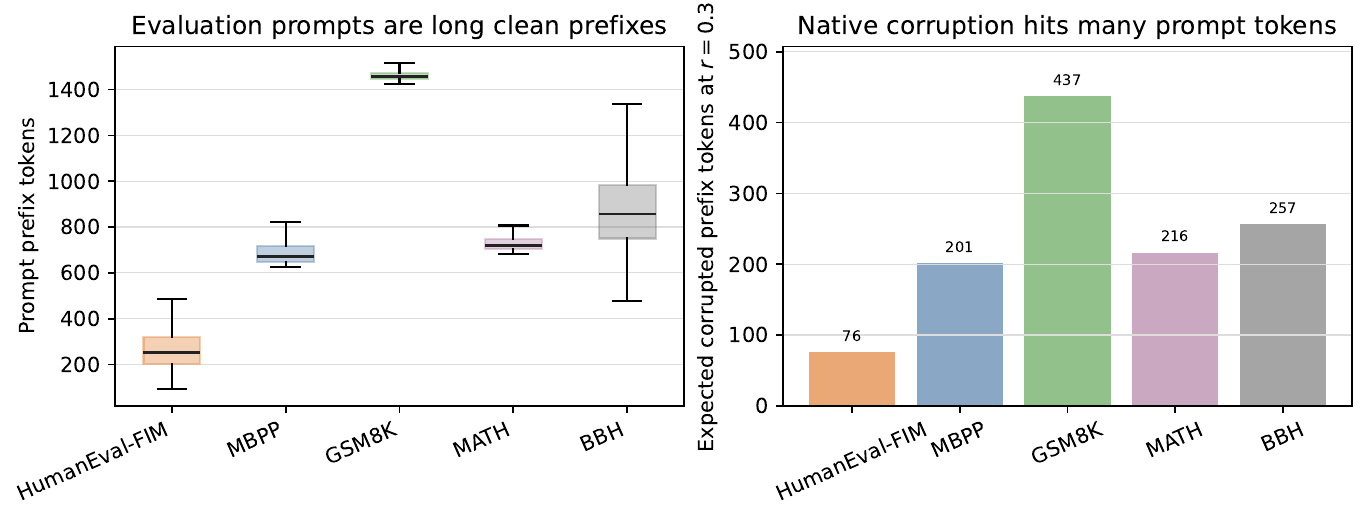}
  \caption{Prompt-length distributions for five benchmarks (left) and expected masked-prefix counts for their median-length prompts at $r=0.3$ (right). HumanEval-FIM uses a deterministic length proxy from HumanEval solutions, rather than model-scoring inputs.}
  \label{fig:benchmark-prefix-mismatch}
\end{figure}

\paragraph{Benchmark and budget dependence.}
At 400B stable, the gains cover all four domains: +2.62 in knowledge/QA, +2.28 in general reasoning, +2.36 in math, and +2.65 in code (Figure~\ref{fig:domain-averages}). The aggregate trend coexists with task-specific variation. For example, MBPP is lower for PCD at 100B (57.60 versus 58.80), but higher at 400B (60.20 versus 59.40). Tables~\ref{tab:main-results} and~\ref{tab:main-coverage} expose the final stable and decay differences.

\paragraph{Clean-prefix structure of benchmark prompts.}
Across the five benchmarks in Figure~\ref{fig:benchmark-prefix-mismatch}, 14,363 public examples rendered with the evaluation templates and a fixed Qwen tokenizer have median prefix and target lengths of 754 and 12 tokens. HumanEval-FIM uses a length proxy. At $r=0.3$, a median-length prefix has 226.2 expected masked tokens, illustrating the clean context available at evaluation.

\paragraph{Evaluation-time prefix corruption.}
In the public Qwen3-1.7B reconstruction study with 50M stable tokens and seeds 42--44, define $\delta(r)=L_{\mathrm{PCD}}-L_{\mathrm{native}}$, where $r$ is the fraction of evaluation prefix tokens masked. The difference changes from $-0.3840\pm0.0013$ nats at $r=0$ to $+0.7573\pm0.1661$ at $r=1$ (mean $\pm$ training-seed SD). The advantage thus reverses when the prefix is fully masked, indicating dependence on visible context rather than robustness to natural prompt errors.

\paragraph{What survives the common decay?}
The stable checkpoint measures objective replacement; matched decay tests what remains after both branches return to block diffusion. Over +50B, native gains +2.57 Avg.6 versus +1.68 for PCD, compressing the gap from +2.56 to +1.67. PCD still leads on average, but this catch-up and two task reversals preclude extrapolation beyond the tested decay.

\begin{figure}[!t]
  \centering
  \begin{minipage}[t]{0.48\textwidth}
    \vspace{0pt}
    \centering
  \begin{tikzpicture}[baseline=(current bounding box.north)]
    \begin{axis}[
      width=\linewidth,
      height=4.2cm,
      title={Transfer under chat SFT},
      title style={font=\small\bfseries},
      xlabel={Chat-SFT epoch},
      ylabel={5-bench Avg.},
      xmin=0.7, xmax=5.3,
      ymin=63.5, ymax=68.4,
      xtick={1,2,3,4,5},
      ytick={64,65,66,67,68},
      grid=major,
      major grid style={gray!18},
      axis x line*=bottom,
      axis y line*=left,
      axis line style={black!75},
      tick label style={font=\footnotesize},
      label style={font=\footnotesize},
      legend style={font=\scriptsize,draw=gray!35,fill=white,fill opacity=0.94,text opacity=1,at={(0.03,0.97)},anchor=north west},
      legend cell align={left},
    ]
      \addplot+[thick,mark=square*,mark size=1.9pt,color=lladablue,mark options={fill=lladablue}] coordinates {
        (1,64.06) (2,65.48) (3,65.12) (4,66.53) (5,65.84)
      };
      \addlegendentry{Native}
      \addplot+[very thick,mark=*,mark size=2.1pt,color=lladaorange,mark options={fill=lladaorange}] coordinates {
        (1,64.72) (2,65.40) (3,67.13) (4,67.07) (5,67.39)
      };
      \addlegendentry{PCD}
      \node[anchor=south,font=\scriptsize\bfseries,text=lladaorange!85!black] at (axis cs:3,67.13) {+2.01};
      \node[anchor=south east,font=\scriptsize\bfseries,text=lladaorange!85!black] at (axis cs:5,67.39) {+1.55};
    \end{axis}
  \end{tikzpicture}
  \caption{Chat-SFT transfer on LLaDA2-Mini. HumanEval-FIM is excluded because chat tuning changes the infilling interface. Points show corresponding epochs from one pair of runs.}
  \label{fig:chat-sft-transfer}
  \end{minipage}\hfill
  \begin{minipage}[t]{0.48\textwidth}
    \vspace{0pt}
    \centering
    \begin{tikzpicture}[baseline=(current bounding box.north)]
      \begin{axis}[
        width=\linewidth,
        height=4.2cm,
        title={AR-interface diagnostic},
        title style={font=\small\bfseries},
        xmin=55.0, xmax=65.2,
        ymin=0.45, ymax=4.60,
        xlabel={AR-decoding Avg.6},
        ylabel={Stable budget},
        ytick={1,2,3,4},
        yticklabels={100B,200B,300B,400B},
        xtick={56,58,60,62,64},
        grid=major,
        major grid style={gray!18},
        axis x line*=bottom,
        axis y line*=left,
        axis line style={black!75},
        tick label style={font=\footnotesize},
        label style={font=\footnotesize},
        legend style={font=\scriptsize,draw=none,at={(0.02,0.98)},anchor=north west,fill=white},
      ]
        \addplot[pcdgray!45,very thick,mark=none,forget plot] coordinates {(60.31,1) (62.10,1)};
        \addplot[pcdgray!45,very thick,mark=none,forget plot] coordinates {(60.72,2) (62.36,2)};
        \addplot[pcdgray!45,very thick,mark=none,forget plot] coordinates {(59.83,3) (62.33,3)};
        \addplot[pcdgray!45,very thick,mark=none,forget plot] coordinates {(60.78,4) (62.61,4)};
        \addplot+[lladablue,only marks,mark=square*,mark options={fill=lladablue,draw=lladablue},mark size=2.2pt] coordinates {
          (60.31,1) (60.72,2) (59.83,3) (60.78,4)
        };
        \addlegendentry{Native}
        \addplot+[lladaorange,only marks,mark=*,mark options={fill=lladaorange,draw=lladaorange},mark size=2.5pt] coordinates {
          (62.10,1) (62.36,2) (62.33,3) (62.61,4)
        };
        \addlegendentry{PCD}
        \node[anchor=west,font=\scriptsize\bfseries,text=lladaorange] at (axis cs:62.10,1) {+1.79};
        \node[anchor=west,font=\scriptsize\bfseries,text=lladaorange] at (axis cs:62.36,2) {+1.64};
        \node[anchor=west,font=\scriptsize\bfseries,text=lladaorange] at (axis cs:62.33,3) {+2.50};
        \node[anchor=west,font=\scriptsize\bfseries,text=lladaorange] at (axis cs:62.61,4) {+1.83};
      \end{axis}
    \end{tikzpicture}
    \caption{AR-interface diagnostic. Avg.6 uses standard HumanEval rather than HumanEval-FIM; orange labels report gains. Both branches use the same evaluation protocol.}
    \label{fig:ar-interface-sanity}
  \end{minipage}
\end{figure}

\paragraph{Transfer under chat SFT.}
We also compare the reported native-derived and PCD-derived chat-SFT runs.
The exact starting checkpoints are not recorded, so this comparison is descriptive and separate from the matched-decay study.
Under the same SFT recipe, the final reported epoch scores 67.39 for the PCD-derived checkpoint versus 65.84 for native SFT, a +1.55-point difference (Figure~\ref{fig:chat-sft-transfer}).
The five corresponding-epoch gaps range from $-0.08$ to $+2.01$, so the advantage is not uniform along the trajectory.
This comparison uses one pair of SFT runs and a shared final epoch, without selecting separate best checkpoints. HumanEval-FIM is excluded because chat tuning changes the infilling interface.

\paragraph{Compatibility with AR decoding.}
Figure~\ref{fig:ar-interface-sanity} evaluates the matched stable checkpoints through their AR interface; Avg.6 replaces HumanEval-FIM with standard HumanEval for causal continuation.
PCD remains above native stable training at all four budgets, with Avg.6 gains of +1.79, +1.64, +2.50, and +1.83 at 100B, 200B, 300B, and 400B, respectively.
These results show that the observed diffusion gains coexist with stronger AR-decoding scores than the matched native baseline.
Both branches use the same MBPP execution timeout of 120 seconds.

\paragraph{What can be attributed to prefix visibility?}
The full-scale comparisons establish a task advantage for the complete recipe, which changes attention, corruption support, and supervision. The separate no-AR intervention isolates training-prefix visibility and finds a reconstruction benefit; evaluation-prefix corruption instead reveals the full recipe's dependence on supplied context. These effects are not an additive explanation of the full-scale task gains. Auxiliary AR's task-level contribution remains unresolved, and the AR-interface comparison uses native stable checkpoints rather than the original AR parent, so it does not establish preservation of the parent's capabilities.

The public generation checks did not meet the paired readiness criterion under the tested samplers. The public controls therefore support reconstruction claims; their effects on generated-answer accuracy and arbitrary infilling remain unestablished.

\paragraph{Comparison unit and uncertainty.}
The 100B--400B points form one paired native/PCD trajectory; matched decay adds 50B tokens to each 400B parent, and chat SFT compares corresponding epochs. The full-scale averages therefore describe completed trajectories rather than independent-training variability. Public reconstruction intervals quantify variation across stable-stage seeds conditional on a shared warmup parent and their declared evaluation interface; they do not estimate uncertainty in full-scale task accuracy.

\FloatBarrier

\section{Conclusion}

PCD constructs clean-prefix conditioning from unlabeled sequences by sampling a boundary and restricting corruption to the suffix. Its full recipe combines suffix denoising with causal prefix supervision.
The treatment of conditioning context during stable training remains consequential in the studied AR-to-diffusion conversion pipeline: PCD's task advantage remains after the matched +50B-token decay studied here, with inference unchanged.
In the primary comparison, PCD improves all 12 tasks at the final stable checkpoint and retains gains on 10 after matched decay.
Separate public studies identify a reconstruction benefit from training-prefix visibility without AR supervision and a context-dependent specialization cost for the full recipe.
These findings support evaluating stable-stage conditioning through both the resulting task behavior and its dependence on visible context. The downstream value of auxiliary AR and the transfer of the isolated visibility effect to full-scale task gains remain unresolved.

\bibliographystyle{antgroup}
\bibliography{custom}

\end{document}